\documentclass[runningheads]{llncs}
\usepackage[T1]{fontenc}
\usepackage{graphicx}
\usepackage{booktabs}
\usepackage{multirow}
\usepackage[misc]{ifsym}
\newcommand{\corr}{(\Letter)}
\usepackage{url} 
\usepackage{mwe}

\begin{document}

\title{Empowering On-Device Model Adaptation with an Edge AI Inference Accelerator}

\titlerunning{Empowering On-Device Model Adaptation with an Edge AI Inf. Accelerator}

\author{Mateusz Piechocki\inst{1} \corr \orcidID{0000-0002-3479-0237} \and
Alessandro Capotondi\inst{2} \orcidID{0000-0001-8705-0761} \and
Marek Kraft\inst{1} \orcidID{0000-0001-6483-2357}}

\authorrunning{M. Piechocki et al.}

\institute{Poznan University of Technology, Institute of Robotics and Machine Intelligence, Poznan, Poland \email{\{name.surname\}@put.poznan.pl}
\and
University of Modena and Reggio Emilia, The High-Performance Real-Time Laboratory, Modena, Italy \email{alessandro.capotondi@unimore.it}
}

\maketitle              

\begin{abstract}

On-device model adaptation is essential to enable lifelong personalization on resource-constrained hardware, but compute, power, and memory limitations of such devices make end-to-end backpropagation impractical for modern deep neural networks. This work proposes a heterogeneous adaptation pipeline that repurposes a commercial edge AI inference accelerator, Hailo-8L, for frozen-backbone feature extraction during on-device training. The computational graph is partitioned so that the pre-trained backbone is quantized to INT8 and run on the accelerator, while only a lightweight FP32 classification head is fine-tuned on the host CPU, enabling frequent, energy-efficient in-field updates with most weights remaining fixed. Across multiple architectures and datasets, this pipeline achieves up to 15.4$\times$ faster wall-clock training time compared to a Raspberry Pi 5 CPU baseline, offers competitive throughput in favorable settings, and consistently reduces energy per sample. Post-training quantization restoration is shown to be crucial for preserving the quality of accelerator-generated features and mitigating accuracy loss in quantization-sensitive architectures. Overall, the results demonstrate a practical approach to efficient on-device adaptation using inference-oriented edge accelerators. The implementation is available at \url{https://github.com/MatPiech/accelerator-training}.

\keywords{on-device training \and model adaptation \and hardware acceleration \and edge AI \and AI accelerator \and edge computing}
\end{abstract}

\section{Introduction}

\begin{figure}[t!]
\centering
\includegraphics[width=\textwidth]{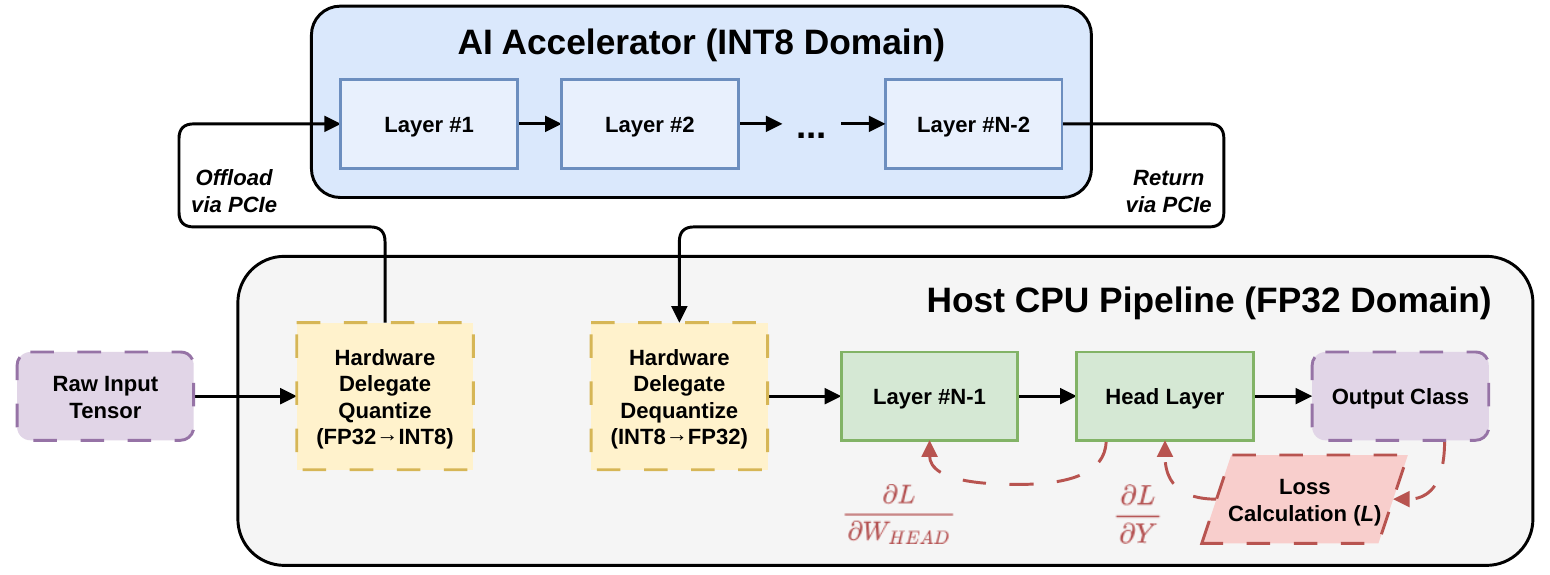}
\caption{Micro-architecture of the heterogeneous on-device adaptation pipeline, detailing graph partitioning, data formats, and gradient flow. Backpropagation is truncated at the accelerator boundary, so only the top adaptive layers receive gradient updates.}
\label{fig:on-device-training-flow}
\end{figure}

Deep learning has become widespread across many application domains, outperforming traditional methods. However, these models typically struggle to generalize to new, unseen data, especially in non-stationary or dynamic environments. To address this, model adaptation techniques have become a key research area, aiming to incrementally update models as conditions change while retaining prior knowledge \cite{10444954,10.1145/3696003}. A fundamental component of effective post-deployment model adaptation is on-device training, which allows a pre-trained model deployed on the target device to adapt to new data, improving predictive performance and enabling personalization to specific users \cite{10.1007/978-3-032-19099-411} or operational contexts \cite{10.1007/978-3-032-07343-328}. Training near the data source reduces reliance on network connectivity and improves privacy, especially for sensitive information. However, it remains challenging due to high computational demands and the complexity of training on resource-constrained hardware \cite{prabhu2023computationally}. Common platforms such as smartphones, edge devices, and embedded systems are limited in processing power, memory, and energy, creating major obstacles for advanced solutions \cite{RAY20221595}. Insufficient hardware often becomes a bottleneck, especially for computationally intensive deep learning applications.

Despite significant theoretical advances in model adaptation, implementing on-device training on resource-constrained hardware remains highly challenging due to limited computational capacity \cite{prabhu2023computationally,Ghunaim_2023_CVPR}. Furthermore, Verwimp et al. \cite{verwimp2024continual} argue that rapid progress in deep learning and its widespread deployment create an urgent need for research on computational efficiency. To address these gaps, this work investigates on-device training using a high-performance, energy-efficient co-processor originally designed to accelerate inference on resource-constrained platforms. By optimizing dataflow and numerical precision, the proposed training pipeline reduces computation time and power consumption, increasing the energy efficiency of on-device learning. This setting reflects a common constraint in lifelong learning: although edge devices often lack the resources required for full backpropagation, they can periodically update a small subset of parameters, such as a task-specific head, as data distributions or user-specific patterns evolve. The proposed adaptation workflow for classification-layer fine-tuning is shown in Fig. \ref{fig:on-device-training-flow}. This study aims to improve the feasibility, energy efficiency, and computational performance of lightweight on-device updates, enabling more flexible model adaptation on edge devices. The main contributions are as follows.

\begin{enumerate}
    \item A heterogeneous ONNX Runtime-based training pipeline that repurposes an INT8 inference accelerator for frozen-backbone feature extraction during on-device head fine-tuning.
    \item A systematic evaluation of post-training quantization restoration strategies and their impact on downstream adaptation performance (accuracy, throughput, and energy).
    \item A system-level comparison against CPU-only and edge-GPU baselines across multiple architectures, datasets, and batch sizes.
\end{enumerate}

\section{Related Work}

Recent advances in machine learning have driven the broad adoption of neural networks within data-processing pipelines, improving downstream performance while increasing overall system complexity. However, conventional episodic training paradigms fail in dynamic resource-constrained settings \cite{11363404}. Consequently, efficient model adaptation strategies have become highly desirable for edge AI deployments, but their use remains limited by the computational cost of training.

In this context, on-device training plays a crucial role in the practical development of self-adaptive, personalized models whose parameters are updated directly on edge platforms \cite{10.1145/3696003}. A leading work in this domain is \cite{NEURIPS2022_90c56c77}, which proposes a \textit{Tiny Training Engine}. This method uses quantization-aware scaling to calibrate gradient magnitudes and stabilize 8-bit quantized training, as well as sparse updates to omit gradient computation for less informative layers and sub-tensors. With these approaches, the full training loop operates within 256 KB SRAM and 1 MB of Flash memory. In terms of hardware accelerators, the authors of \cite{das2022enabling} use a smartphone GPU to train a classification network using OpenCL-optimized matrix multiplication kernels. However, full training remains slower than on CPUs due to data-movement bottlenecks. In contrast, authors in \cite{11130214} present an end-to-end, hardware-accelerated on-device learning strategy that uses robustness analysis and energy estimation to partition neural networks into frozen INT8 and adaptive Bfloat16 segments, specifically targeting deployment on specialized edge accelerators. While the reported results are promising, the evaluation is limited to the VGG-16 model trained on the CIFAR-10 dataset.

Despite these advances, conventional edge devices remain insufficient to support efficient on-device training, underscoring the need for alternative computational paradigms. While several notable studies exist and specialized co-processors for deep learning inference have been widely deployed, the current body of literature lacks investigations that exploit such inference co-processors to accelerate the training process itself on edge platforms.

\section{Experimental Methodology}

\subsection{Heterogeneous On-Device Adaptation Architecture}

The core of the proposed methodology relies on an asymmetric execution pipeline designed specifically for resource-constrained edge devices. The workflow fundamentally divides the learning process into two distinct hardware domains: an INT8 hardware accelerator for high-throughput feature extraction and a FP32 host CPU for adaptive optimization.

As illustrated in Fig. \ref{fig:on-device-training-flow}, the model backbone parameters are offloaded to the hardware accelerator to generate feature maps during the forward pass but are excluded from the backward pass. Specifically, the backpropagation procedure is mathematically truncated prior to the accelerator, so gradient computations and parameter updates occur only in the adaptive top layers on the host CPU. This confines expensive training operations to the CPU and reduces on-device training latency, while most inference runs on the high-throughput accelerator. Formally, each model is split into a frozen feature extractor and a trainable classification head:

\begin{equation}
f(x; \theta_b, \theta_h) = h\bigl(g(x; \theta_b); \theta_h\bigr)
\end{equation}

where $g(\cdot; \theta_b)$ denotes the pre-trained backbone and $h(\cdot; \theta_h)$ represents the task-specific classification head. During on-device adaptation, the backbone parameters $\theta_b$ are frozen and only $\theta_h$ is updated by minimizing the loss on the target dataset. Consequently, no gradients are computed for the accelerator-executed backbone, and the backward pass stops at the boundary between the accelerator-generated feature tensor and the host-side trainable head.

\subsection{Model Conversion, Quantization, and Performance Restoration}

\begin{figure}[t!]
\centering
\includegraphics[width=\textwidth]{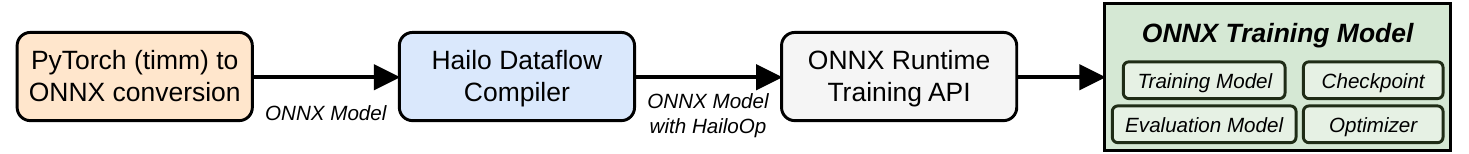}
\caption{The proposed ONNX-based toolchain for Hailo-accelerated adaptation. PyTorch models are exported to ONNX, then the Hailo Dataflow Compiler applies 8-bit quantization and restoration to generate the hardware delegate (\textit{HailoOp}). The ONNX Runtime Training API then processes this hybrid graph to produce training artifacts.}
\label{fig:model-conversion}
\end{figure}

While PyTorch offers substantial flexibility for model development, its dynamic execution graph is inherently suboptimal for deployment and training on resource-constrained edge devices. Consequently, this study leverages the ONNX Runtime framework, specifically its training submodule, due to its support for a wide range of hardware accelerators, cross-platform compatibility, and versatility.

The conversion and compilation workflow (Fig. \ref{fig:model-conversion}) begins by exporting standard pre-trained models to ONNX. The Hailo Dataflow Compiler \cite{hailo-dataflow} then processes the frozen backbone, quantizes it to 8-bit integers, applies a performance restoration strategy, and embeds a hardware delegate node (\textit{HailoOp}) with the compiled backbone into the ONNX model. The ONNX Runtime Training API runs on this hybrid computation graph to generate four artifacts for on-device training: (i) a training graph with the forward pass and truncated backward pass; (ii) an evaluation graph for inference; (iii) a checkpoint file with the model’s trainable parameters; and (iv) an optimizer file defining the optimizer state.

\begin{table}[t!]
\centering
\setlength{\tabcolsep}{6pt}
\renewcommand{\arraystretch}{1.15}
\caption{Evaluated performance restoration strategies and their configurations. Eq. denotes channel equalization, IBC represents iterative bias correction, FT denotes knowledge-distillation fine-tuning, and AR stands for the AdaRound algorithm.}
\label{tab:perf-restoration-strategies}
\begin{tabular}{c c c l}
\toprule
\textbf{\#} & \textbf{Techniques} & \textbf{Cost} & \textbf{Configuration Details} \\
\midrule
0 & None & -- & -- \\
1 & Eq. & low & -- \\
2 & Eq. + IBC & low & 64 unlabeled images \\
3 & Eq. + FT & medium & 4 epochs \& 1024 images \\
4 & Eq. + AR & high & 320 epochs \& 256 images on all layers \\
\bottomrule
\end{tabular}
\end{table}

The deployment of models on resource-limited edge co-processors requires converting floating-point parameters to low-precision formats, typically INT8. However, naive post-training quantization can severely distort intermediate feature maps, degrading upstream gradients during host-side adaptation. To preserve gradient stability, various performance restoration strategies were evaluated to recover the integrity and predictive performance of quantized models. As summarized in Table \ref{tab:perf-restoration-strategies}, five strategies were evaluated, ranging from conventional channel equalization (Eq.) \cite{meller2019same} to more computationally demanding, dataset-calibrated methods: iterative bias correction (IBC) \cite{finkelstein2019fighting}, knowledge-distillation-based fine-tuning (FT) \cite{10.1007/978-3-031-25082-88}, and AdaRound (AR) \cite{nagel2020up}. Eq. is a general, low-cost method that addresses scale imbalances by inversely rescaling weights of neighboring layers and thus serves as a baseline applicable to all configurations. Building on this, IBC applies a low-cost bias calibration to correct mean shifts. FT jointly optimizes the quantized model’s weights, scales, and biases via knowledge distillation to address parameter desynchronization. Finally, AR reduces rounding errors via a data-adapted weight-rounding scheme that minimizes local task loss. All methods were implemented with the Hailo Dataflow Compiler using the default configuration without additional tuning. The compilation was performed on a workstation with an NVIDIA GeForce RTX 4090 GPU and an AMD Ryzen Threadripper PRO 5955WX CPU. To support this process, a calibration set of 1024 images was randomly sampled from the \textit{ILSVRC2012\_val} split of ImageNet \cite{deng2009imagenet}.

\subsection{Experimental Setup}

\noindent\textbf{Models.} To validate the architecture-agnostic design of the proposed pipeline, four diverse architectures were selected from the \textit{timm} collection \cite{rw2019timm}: ResNet18 \cite{7780459ResNet}, EfficientNet-Lite 4 \cite{pmlr-v97-tan19a}, MobileNetV3 Large \cite{howard2019searching}, and FastViT-SA12 \cite{10377971}. Table \ref{tab:model-summary} summarizes the computational characteristics of the frozen backbones.

\noindent\textbf{Datasets.} The system's adaptability was evaluated on two downstream image classification datasets: CIFAR-100 \cite{Krizhevsky09learningmultiple} and Oxford-IIIT Pet \cite{parkhi12a}. CIFAR-100 contains 60,000 color images, uniformly distributed over 100 classes, while Oxford-IIIT Pet comprises images from 37 dog and cat breeds with about 200 samples per category. For consistency, all images were resized to a spatial resolution of $224\times224$ and normalized using ImageNet statistics.

\begin{table}[t!]
\centering
\setlength{\tabcolsep}{0.5em}
\renewcommand{\arraystretch}{1.1}
\caption{Frozen-backbone characterization at a $3\times224\times224$ input resolution. Metrics reflect the computational complexity delegated to the hardware accelerator, explicitly excluding the adaptive, host-side classification head. $B$ denotes batch size.}
\label{tab:model-summary}
\begin{tabular}{lrrr}
\toprule
\textbf{Model} & \textbf{MACs (G)} & \textbf{Params (M)} & \textbf{Output Features} \\
\midrule
ResNet18              & 1.82 & 11.19 & $B \times 512$ \\
EfficientNet-Lite 4   & 1.35 & 11.67 & $B \times 1280$ \\
MobileNetV3 Large     & 0.22 &  4.23 & $B \times 1280$ \\
FastViT-SA12          & 1.38 & 10.55 & $B \times 1024$ \\
\bottomrule
\end{tabular}
\end{table}

\noindent\textbf{Hardware platforms.} The experimental evaluation used the Hailo-8L edge AI accelerator \cite{hailo-8l}, a high-performance co-processor delivering about 13 Tera Operations Per Second (TOPS) and a power consumption of 1.5~W. In the heterogeneous accelerated on-device training pipeline, a Hailo-8L module was interfaced with a Raspberry Pi 5 (RPi 5), a low-cost, fanless, compact single-board computer with a Broadcom BCM2712 system-on-chip, featuring a quad-core Arm Cortex-A76 CPU operating at 2.4~GHz and 4~GB of RAM. The connection used an M.2 HAT expansion board, which provides the physical and electrical interface between the co-processor and the host. For system-level benchmarking, this heterogeneous setup was compared with CPU-only computation on the RPi 5 and with a high-performance, energy-efficient edge GPU platform, the NVIDIA Jetson Orin Nano \cite{orinnano}, which integrates an NVIDIA Ampere GPU with 1024 CUDA cores and 32 Tensor Cores, delivering up to 40 TOPS at 15~W power mode.

\noindent\textbf{Evaluation protocol.} For a fair comparison, all models follow the same adaptation protocol -- the pre-trained backbone is frozen and only the classification head is fine-tuned. This head-only update reflects typical in-field personalization and serves as a building block for parameter-efficient fine-tuning methods that restrict gradient updates to small add-on modules while keeping the backbone mostly fixed. In the RPi 5 CPU baseline, both the frozen backbone and trainable head use FP32 data format. In the Jetson Orin Nano baseline, the same head-only fine-tuning runs in FP32 on the embedded GPU. In the proposed heterogeneous configuration, the frozen backbone is quantized to INT8 and operates on the Hailo-8L accelerator, while the classification head, loss computation, backward pass, and optimizer step run in FP32 on the RPi 5 CPU. All configurations used ONNX Runtime v1.22.1 \cite{custom-onnxruntime} with the Hailo execution provider (HailoRT v4.23.0) \cite{hailo} integrated specifically to support the proposed heterogeneous workflow. All experiments used the cross-entropy loss and the AdamW optimizer with a fixed learning rate of $1\times10^{-3}$. To assess the impact of batched execution, training batch sizes of 1, 4, and 16 were used, while the evaluation always used a batch size of 1. Model performance is reported as test-set classification accuracy after five epochs, with a fixed random seed employed for reproducibility.

\noindent\textbf{Energy measurement.} For each configuration, power consumption was measured for a single training epoch under the same workload (end-to-end head fine-tuning, including feature extraction, loss, backpropagation, and optimizer step) and reported without subtracting idle power. On the Jetson Orin Nano, on-module power sensors were read with \texttt{tegrastats} at 1~Hz. On the RPi 5, for both the CPU baseline and the accelerator-based architecture, an external inline power meter was employed, sampling at 5~Hz.

\section{Results}

\subsection{System-Level Profiling}

\begin{table}[t!]
\caption{Experimental evaluation of end-to-end training performance, throughput, and energy efficiency with batch size 1. The proposed heterogeneous pipeline (Hailo-8L + RPi 5) is compared with pure-host (RPi 5) and edge GPU (Orin Nano) baselines. Numbers in parentheses indicate the best performance-restoration strategy per dataset, for which results are reported. In the table, \textit{Acc.} means test accuracy (\%), \textit{Tput} is training throughput (ms/sample), and \textit{E} stands for training energy consumption (mJ/sample). Bold values mark the best result for each model in that column.}
\label{tab:high-level-results}
\setlength{\tabcolsep}{0.4em}
\def\arraystretch{1.1}
\centering
\resizebox{\textwidth}{!}{%
\begin{tabular}{llcccccc}
\hline\noalign{\smallskip}
\multirow{2}{*}{\textbf{Model}} & \multirow{2}{*}{\textbf{Hardware}} & \multicolumn{3}{c}{\textbf{CIFAR-100}} & \multicolumn{3}{c}{\textbf{Oxford-IIIT Pet}} \\
\cmidrule(lr){3-5}\cmidrule(lr){6-8}
 &  & \textbf{Acc.} & \textbf{Tput} & \textbf{E} & \textbf{Acc.} & \textbf{Tput} & \textbf{E} \\
\hline
\multirow{3}{*}{ResNet18} & RPi 5 (CPU) & \textbf{64.68} & 92.85 & 525.65 & 86.97 & 102.84 & 561.25 \\
 & Orin Nano (GPU) & 64.67 & 13.70 & 128.25 & \textbf{87.00} & 9.22 & 128.13 \\
 & \textbf{Proposed (\#4, \#4)} & 64.62 & \textbf{6.04} & \textbf{38.65} & 86.94 & \textbf{7.17} & \textbf{46.35} \\
\hline
\multirow{3}{*}{\shortstack{EfficientNet\\-Lite 4}} & RPi 5 (CPU) & 58.48 & 178.28 & 1042.91 & 83.37 & 194.02 & 1064.18 \\
 & Orin Nano (GPU) & 58.80 & \textbf{23.56} & 234.09 & 83.43 & 24.39 & 184.00 \\
 & \textbf{Proposed (\#3, \#4)} & \textbf{60.23} & 23.91 & \textbf{124.64} & \textbf{84.08} & \textbf{23.01} & \textbf{111.99} \\
\hline
\multirow{3}{*}{\shortstack{MobileNetV3\\Large}} & RPi 5 (CPU) & \textbf{68.26} & 57.33 & 286.97 & \textbf{88.88} & 52.71 & 315.63 \\
 & Orin Nano (GPU) & 68.10 & \textbf{9.49} & 110.04 & 88.83 & \textbf{9.30} & 102.19 \\
 & \textbf{Proposed (\#4, \#4)} & 55.53 & 13.73 & \textbf{78.22} & 87.00 & 12.31 & \textbf{65.37} \\
\hline
\multirow{3}{*}{FastViT-SA12} & RPi 5 (CPU) & 71.19 & 223.89 & 1224.60 & \textbf{91.88} & 229.26 & 1228.60 \\
 & Orin Nano (GPU) & \textbf{71.20} & \textbf{21.92} & 306.73 & \textbf{91.88} & 27.31 & 220.37 \\
 & \textbf{Proposed (\#3, \#3)} & 50.19 & 22.81 & \textbf{119.64} & 83.54 & \textbf{21.84} & \textbf{105.80} \\
\hline
\end{tabular}%
}
\end{table}

\begin{figure}[t!]
\centering
\includegraphics[width=\textwidth]{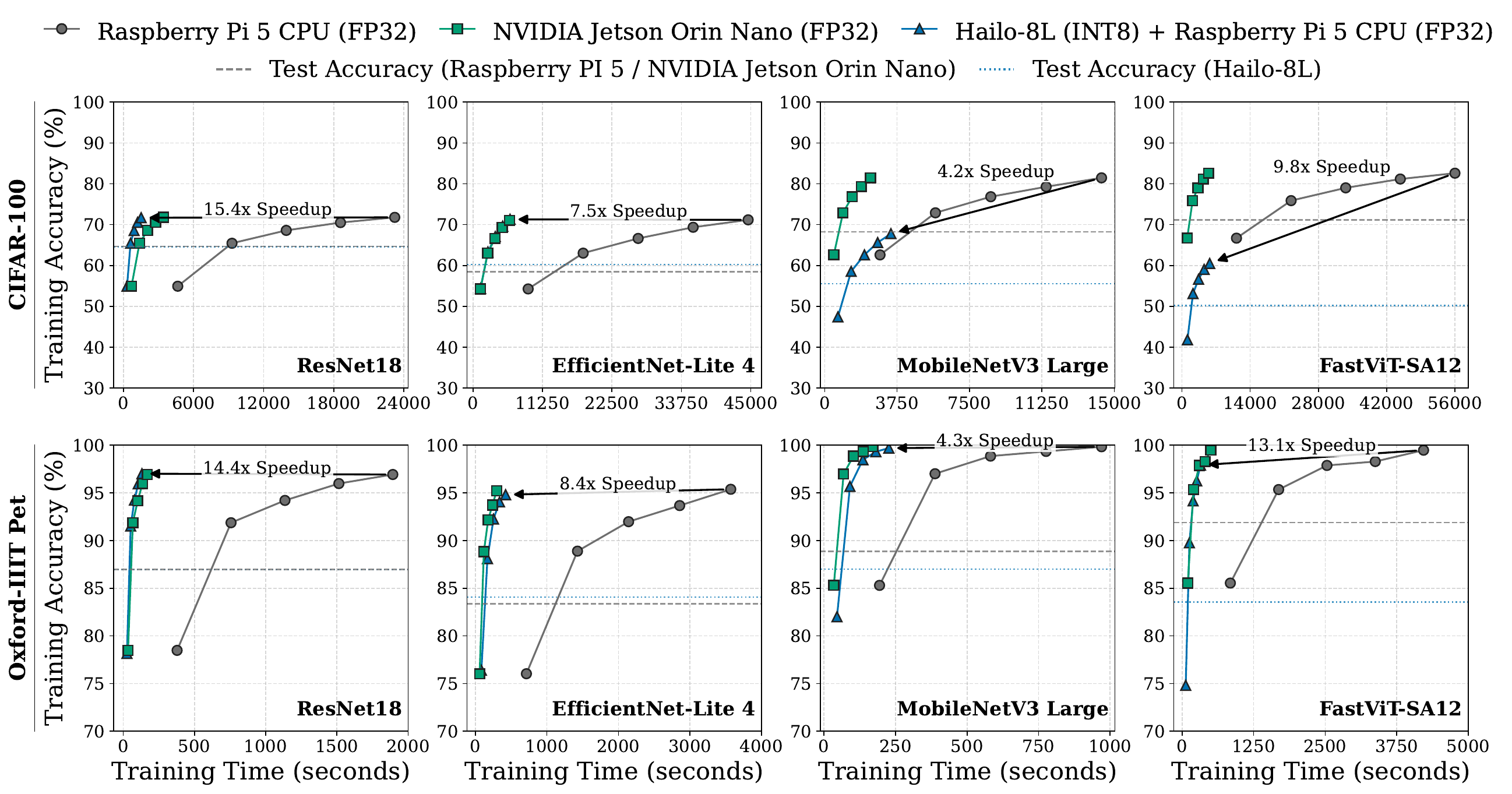}
\caption{Time-to-accuracy convergence profiles for the evaluated architectures and datasets. Curves denote training accuracy over time; markers indicate the values for subsequent training epochs; horizontal dashed lines represent final test accuracy.}
\label{fig:convergence}
\end{figure}

The primary objective of the proposed heterogeneous pipeline is to accelerate on-device model training. As shown in the time-to-accuracy convergence profiles (Fig. \ref{fig:convergence}), combining the Hailo-8L accelerator with the Raspberry Pi 5 outperforms the CPU-only baseline. For convolutional models such as ResNet18, it reduces wall-clock training time by up to 15.4$\times$, while achieving test accuracy close to the unquantized FP32 reference. As summarized in Table \ref{tab:high-level-results}, this setup reaches 6.04 ms/sample for ResNet18, which is not only significantly better than the CPU-only setup, but for this architecture also outperforms the dedicated edge GPU configuration (NVIDIA Jetson Orin Nano at 13.70 ms/sample), while using 3.3$\times$ less energy per trained sample.

Nevertheless, it is important to note the variation in acceleration observed across different architectures. Although ResNet18 exceeds a $15\times$ speedup, MobileNetV3 Large exhibits a comparatively modest acceleration of approximately $4\times$. This gap stems mainly from network design: MobileNetV3 employs a computationally intensive, high-parameter classifier head that executes on the non-accelerated CPU, making the host-side FP32 backward pass the primary latency bottleneck.

\subsection{Impact of Quantization and Performance Restoration Methods}

\begin{figure}[t!]
\centering
\includegraphics[width=\textwidth]{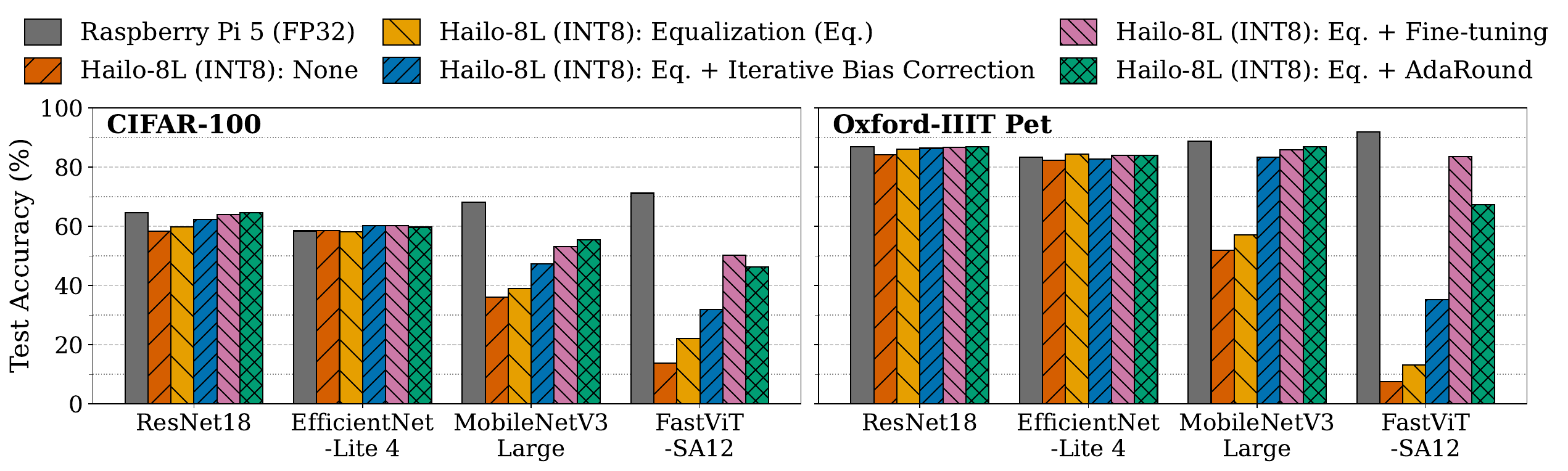}
\caption{Impact of INT8 post-training quantization and subsequent performance recovery strategies on upstream gradient stability and final test accuracy, assessed following five epochs of on-device training on the CIFAR-100 and Oxford-IIIT Pet datasets.}
\label{fig:quantization-performance-restoration}
\end{figure}

Although hardware-level partitioning yields notable computational acceleration, asymmetric precision mapping simultaneously introduces a vulnerability to distortion of upstream gradients when the INT8 feature representations are heavily degraded. Figure \ref{fig:quantization-performance-restoration} illustrates the effect of post-training quantization and evaluates the effectiveness of performance restoration strategies.

The empirical results reveal a clear architectural divide in quantization resilience. Without performance restoration, direct INT8 post-training quantization can substantially reduce accuracy relative to FP32 baselines. Quantization-resilient backbones, such as ResNet18 and EfficientNet-Lite, incur only moderate drops, and lightweight channel equalization or iterative bias correction (configurations \#1 and \#2 in Table \ref{tab:perf-restoration-strategies}) typically suffices to recover baseline accuracy. In contrast, architectures using Hard-Swish (MobileNetV3) or self-attention (FastViT-SA12) are more vulnerable to INT8 degradation. For these models, simple equalization is insufficient, and stronger restoration strategies are required; even then, the remaining accuracy gap can depend on the dataset. As shown in Fig. \ref{fig:quantization-performance-restoration}, such models require more advanced data-driven methods, especially knowledge-distillation-based fine-tuning or AdaRound (configurations \#3 and \#4), to maintain intermediate representation fidelity and keep test accuracy of the heterogeneous quantized pipeline close to the FP32 reference.

Based on these findings, a cost-efficient strategy is to start with low-cost restoration methods (\#1–\#2) for standard convolutional backbones with ReLU or SiLU activations (e.g., ResNet18, EfficientNet-Lite), and move to more complex data-driven methods (\#3–\#4) only if accuracy is still insufficient. This is especially important for backbones more sensitive to quantization (e.g., MobileNetV3, FastViT-SA12). With a limited budget for performance restoration, knowledge-distillation fine-tuning (\#3) offers a good accuracy–cost trade-off, while AdaRound (\#4) should be used only when the remaining accuracy degradation is unacceptable.

\subsection{Operational Throughput and Batch Execution Scaling}

\begin{figure}[t!]
\centering
\includegraphics[width=\textwidth]{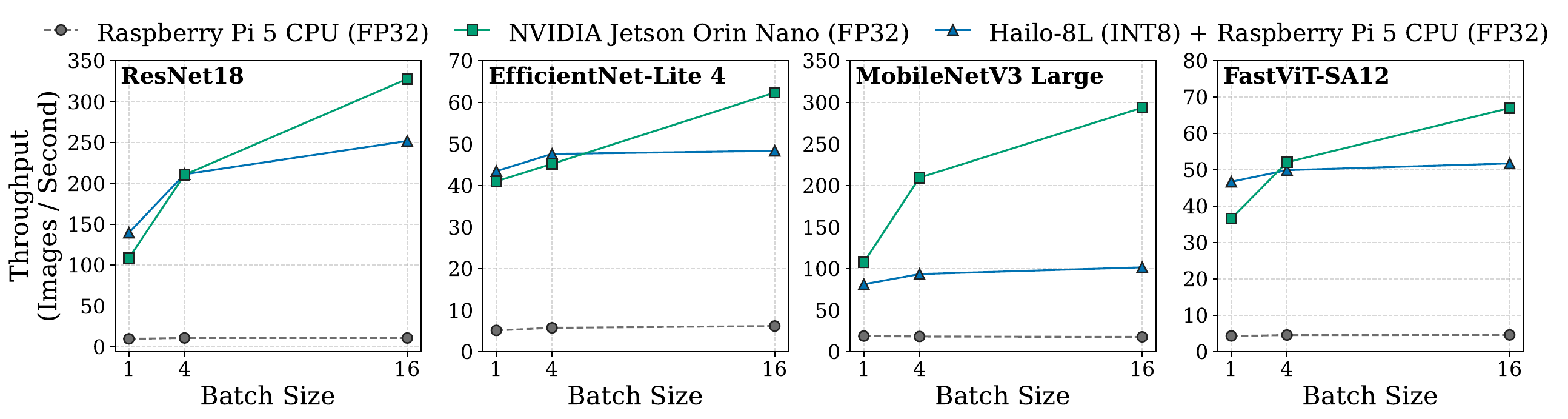}
\caption{Empirical training throughput on the Oxford-IIIT Pet dataset, measured in images per second, across varying batch sizes (1, 4, and 16).}
\label{fig:throughput}
\end{figure}

Beyond convergence time, the evaluation of raw operational throughput (images per second) reveals how the hardware handles execution bottlenecks and exploits data-level parallelism. Figure \ref{fig:throughput} reports the empirically observed training throughput for the examined architectures at batch sizes of 1, 4, and 16.

The pure-host RPi 5 CPU baseline shows a clear computational bottleneck. Across all four network topologies, CPU-only throughput remains low and barely scales with batch size. In contrast, the proposed heterogeneous pipeline removes this bottleneck by offloading feature extraction to the accelerator and using the host CPU only for the adaptive head. This enables good scalability for small batches (1 and 4) and near-saturation throughput for batch sizes above 4. Furthermore, the throughput analysis also shows a clear advantage over a dedicated edge GPU at low batch sizes. For networks such as ResNet-18, EfficientNet-Lite 4 and FastViT-SA12, the Hailo-8L configuration matches or exceeds Orin Nano throughput at batch sizes 1 and 4, while consuming much less power. This is due to its inference-centric architectural design, which underpins its strong performance at small batch sizes, making it well suited to tightly resource-constrained on-device adaptation scenarios where large batches are undesirable.

\section{Conclusion}

This work demonstrates that repurposing inference accelerators can overcome key computational bottlenecks in on-device training. By strictly partitioning the execution graph and offloading the frozen feature extractor to the Hailo-8L hardware accelerator, training latency was reduced and energy efficiency was improved, while competitive throughput was observed for favorable architectures. Empirical results also show that performance restoration techniques, typically used for inference models before deployment, are similarly effective in a training context to stabilize upstream gradients and mitigate accuracy degradation.

From an on-device learning perspective, the main takeaway is that the proposed heterogeneous architecture enables frequent, small-scale model updates under strict energy and latency constraints. This setting is relevant to many edge scenarios where the core perceptual backbone stays fixed while label semantics evolve. Examples include human activity recognition, where a deployed model is periodically personalized to a user via head or adapter updates; local environmental sensing, where on-device adaptation tracks evolving site-specific microclimates; and vision-based industrial defect detection or classification.

At the same time, this study is limited to frozen-backbone fine-tuning, whose effectiveness depends on architecture-specific robustness to INT8 feature degradation. Host-side head computation and cross-domain overheads (quantization, dequantization, and host-accelerator data transfers) can also limit achievable speedups, particularly for models with complex heads or relatively small backbones. Peak memory usage was not quantified in this study; instead, the evaluation was limited to latency/throughput and energy metrics. Finally, the evaluation is restricted to supervised image classification adaptation on a limited set of datasets and model architectures. Future work will extend the heterogeneous framework with parameter-efficient fine-tuning paradigms and validate it on edge-AI workloads in dynamic continual learning, test-time adaptation, and streaming scenarios, using methods to mitigate catastrophic forgetting (e.g., rehearsal, drift detection, selective unfreezing) beyond head-only updates.

\begin{credits}
\subsubsection{\ackname} This research was supported by the European Union’s Horizon Europe Research and Innovation Programme under the project dAIEDGE (Grant Agreement No. 101120726), and by the dAIEDGE Open Call 1, under Sub-grant Agreement No. dAI1OC01. The APC was also funded by Poznan University of Technology, internal grant 0214/SBAD/0253.

\subsubsection{\discintname} The authors have no competing interests to declare that are relevant to the content of this article.

\end{credits}
%
%
%
\bibliographystyle{splncs04}
\bibliography{mybibliography}

@ARTICLE{10444954,
  author  = {Wang, Liyuan and Zhang, Xingxing and Su, Hang and Zhu, Jun},
  journal = {IEEE Transactions on Pattern Analysis and Machine Intelligence},
  title   = {A Comprehensive Survey of Continual Learning: Theory, Method and Application},
  year    = {2024},
  pages   = {1--20},
  doi     = {10.1109/TPAMI.2024.3367329}
}

@article{10.1145/3696003,
author  = {Zhu, Shuai and Voigt, Thiemo and Rahimian, Fatemeh and Ko, Jeonggil},
title   = {On-device Training: A First Overview on Existing Systems},
journal = {ACM Trans. Sen. Netw.},
year    = {2024},
volume  = {20},
number  = {6},
articleno = {118},
doi     = {10.1145/3696003}
}

@inproceedings{prabhu2023computationally,
  author    = {Prabhu, Ameya and Al Kader Hammoud, Hasan Abed and Dokania, Puneet and Torr, Philip H.S. and Lim, Ser-Nam and Ghanem, Bernard and Bibi, Adel},
  title     = {Computationally Budgeted Continual Learning: What Does Matter?},
  booktitle = {2023 IEEE/CVF Conference on Computer Vision and Pattern Recognition (CVPR)},
  year      = {2023},
  pages     = {3698--3707},
  doi       = {10.1109/CVPR52729.2023.00360}
}

@article{RAY20221595,
  author  = {Ray, Partha Pratim},
  title   = {A review on TinyML: State-of-the-art and prospects},
  journal = {Journal of King Saud University - Computer and Information Sciences},
  year    = {2022},
  volume  = {34},
  number  = {4},
  pages   = {1595--1623},
  doi     = {10.1016/j.jksuci.2021.11.019}
}

@InProceedings{Ghunaim_2023_CVPR,
  author    = {Ghunaim, Yasir and Bibi, Adel and Alhamoud, Kumail and Alfarra, Motasem and Hammoud, Hasan Abed Al Kader and Prabhu, Ameya and Torr, Philip H.S. and Ghanem, Bernard},
  title     = {Real-Time Evaluation in Online Continual Learning: A New Hope},
  booktitle = {2023 IEEE/CVF Conference on Computer Vision and Pattern Recognition (CVPR)},
  year      = {2023},
  pages     = {11888--11897},
  doi       = {10.1109/CVPR52729.2023.01144}
}

@article{verwimp2024continual,
  author  = {Verwimp, Eli and Ben-David, Shai and Bethge, Matthias and Cossu, Andrea and Gepperth, Alexander and Hayes, Tyler L. and H{\"u}llermeier, Eyke and Kanan, Christopher and Kudithipudi, Dhireesha and Lampert, Christoph H. and others},
  title   = {Continual Learning: Applications and the Road Forward},
  journal = {Transactions on Machine Learning Research},
  year    = {2024}
}

@InProceedings{10.1007/978-3-032-19099-411,
  author    = {Karag{\"u}l, Muhammed Talha and Yozgat, Ensar Muhammet and As{\i}ll{\i}o{\u{g}}lu, Feyzullah and Arslan, Sanem},
  title     = {On-Device Learning for Human Activity Recognition on~Low-Power Microcontrollers},
  booktitle = {Machine Learning and Principles and Practice of Knowledge Discovery in Databases},
  year      = {2026},
  pages     = {148--160},
  doi       = {10.1007/978-3-032-19099-4_11}
}

@InProceedings{10.1007/978-3-032-07343-328,
  author    = {Piechocki, Mateusz and Kraft, Marek and Capotondi, Alessandro},
  title     = {On-Device Continual Adaptation for~Reliable Solar Irradiance Forecasting},
  booktitle = {Advanced Concepts for Intelligent Vision Systems},
  year      = {2026},
  pages     = {353--364},
  doi       = {10.1007/978-3-032-07343-3_28}
}

@inproceedings{NEURIPS2022_90c56c77,
  author    = {Lin, Ji and Zhu, Ligeng and Chen, Wei-Ming and Wang, Wei-Chen and Gan, Chuang and Han, Song},
  title     = {On-device training under 256KB memory},
  booktitle = {Proceedings of the 36th International Conference on Neural Information Processing Systems},
  year      = {2022}
}

@inproceedings{das2022enabling,
  author    = {Das, Anish and Kwon, Young D. and Chauhan, Jagmohan and Mascolo, Cecilia},
  title     = {Enabling On-Device Smartphone GPU based Training: Lessons Learned},
  booktitle = {2022 IEEE International Conference on Pervasive Computing and Communications Workshops and other Affiliated Events (PerCom Workshops)},
  year      = {2022},
  pages     = {533--538},
  doi       = {10.1109/PerComWorkshops53856.2022.9767442}
}

@INPROCEEDINGS{11130214,
  author    = {Topko, Iuliia and Serdyuk, Alexey and Harbaum, Tanja and Becker, Jürgen},
  title     = {Hardware-Accelerated On-Device Learning: Training, Partitioning, and Compilation for Constrained Edge AI},
  booktitle = {2025 IEEE Computer Society Annual Symposium on VLSI (ISVLSI)},
  year      = {2025},
  volume    = {1},
  pages     = {1--6},
  doi       = {10.1109/ISVLSI65124.2025.11130214}
}

@ARTICLE{11363404,
  author  = {Wu, Beining and Ding, Zihao and Huang, Jun},
  journal = {IEEE Transactions on Network Science and Engineering},
  title   = {A Review of Continual Learning in Edge AI},
  year    = {2026},
  volume  = {13},
  pages   = {6571--6588},
  doi     = {10.1109/TNSE.2026.3657652}
}

@inproceedings{deng2009imagenet,
  author    = {Deng, Jia and Dong, Wei and Socher, Richard and Li, Li-Jia and Li, Kai and Li, Fei-Fei},
  title     = {ImageNet: A large-scale hierarchical image database},
  booktitle = {2009 IEEE Conference on Computer Vision and Pattern Recognition},
  year      = {2009},
  pages     = {248--255},
  doi       = {10.1109/CVPR.2009.5206848}
}

@TECHREPORT{Krizhevsky09learningmultiple,
    author = {Alex Krizhevsky},
    title = {Learning multiple layers of features from tiny images},
    institution={University of Toronto},
    year = {2009}
}

@InProceedings{parkhi12a,
  author    = {Parkhi, Omkar M. and Vedaldi, Andrea and Zisserman, Andrew and Jawahar, C. V.},
  title     = {Cats and dogs},
  booktitle = {2012 IEEE Conference on Computer Vision and Pattern Recognition},
  year      = {2012},
  pages     = {3498--3505},
  doi       = {10.1109/CVPR.2012.6248092}
}

@misc{rw2019timm,
  author = {Ross Wightman},
  title = {PyTorch Image Models},
  year = {2019},
  publisher = {GitHub},
  journal = {GitHub repository},
  doi = {10.5281/zenodo.4414861},
}

@INPROCEEDINGS{howard2019searching,
  author    = {Howard, Andrew and Sandler, Mark and Chen, Bo and Wang, Weijun and Chen, Liang-Chieh and Tan, Mingxing and Chu, Grace and Vasudevan, Vijay and Zhu, Yukun and Pang, Ruoming and Adam, Hartwig and Le, Quoc},
  title     = {Searching for MobileNetV3},
  booktitle = {2019 IEEE/CVF International Conference on Computer Vision (ICCV)},
  year      = {2019},
  pages     = {1314--1324},
  doi       = {10.1109/ICCV.2019.00140}
}

@INPROCEEDINGS{7780459ResNet,
  author    = {He, Kaiming and Zhang, Xiangyu and Ren, Shaoqing and Sun, Jian},
  title     = {Deep Residual Learning for Image Recognition},
  booktitle = {2016 IEEE Conference on Computer Vision and Pattern Recognition (CVPR)},
  year      = {2016},
  pages     = {770--778},
  doi       = {10.1109/CVPR.2016.90}
}

@INPROCEEDINGS{10377971,
  author    = {Anasosalu Vasu, Pavan Kumar and Gabriel, James and Zhu, Jeff and Tuzel, Oncel and Ranjan, Anurag},
  title     = {FastViT: A Fast Hybrid Vision Transformer using Structural Reparameterization},
  booktitle = {2023 IEEE/CVF International Conference on Computer Vision (ICCV)},
  year      = {2023},
  pages     = {5762--5772},
  doi       = {10.1109/ICCV51070.2023.00532}
}

@InProceedings{pmlr-v97-tan19a,
  title = 	 {{E}fficient{N}et: Rethinking Model Scaling for Convolutional Neural Networks},
  author =       {Tan, Mingxing and Le, Quoc},
  booktitle = 	 {Proceedings of the 36th International Conference on Machine Learning},
  pages = 	 {6105--6114},
  year = 	 {2019},
  editor = 	 {Chaudhuri, Kamalika and Salakhutdinov, Ruslan},
  volume = 	 {97},
  series = 	 {Proceedings of Machine Learning Research},
  month = 	 {09--15 Jun},
  publisher =    {PMLR},
  url = 	 {https://proceedings.mlr.press/v97/tan19a.html},
}

@inproceedings{meller2019same,
  title = 	 {Same, Same But Different: Recovering Neural Network Quantization Error Through Weight Factorization},
  author =       {Meller, Eldad and Finkelstein, Alexander and Almog, Uri and Grobman, Mark},
  booktitle = 	 {Proceedings of the 36th International Conference on Machine Learning},
  pages = 	 {4486--4495},
  year = 	 {2019},
  editor = 	 {Chaudhuri, Kamalika and Salakhutdinov, Ruslan},
  volume = 	 {97},
  series = 	 {Proceedings of Machine Learning Research},
  month = 	 {09--15 Jun},
  publisher =    {PMLR},
  url = 	 {https://proceedings.mlr.press/v97/meller19a.html},
}

@article{finkelstein2019fighting,
  title={Fighting quantization bias with bias},
  author={Finkelstein, Alexander and Almog, Uri and Grobman, Mark},
  journal={arXiv preprint arXiv:1906.03193},
  year={2019}
}

@InProceedings{10.1007/978-3-031-25082-88,
  author    = {Finkelstein, Alex and Fuchs, Ella and Tal, Idan and Grobman, Mark and Vosco, Niv and Meller, Eldad},
  title     = {QFT: Post-training Quantization via Fast Joint Finetuning of All Degrees of Freedom},
  booktitle = {Computer Vision -- ECCV 2022 Workshops},
  year      = {2023},
  pages     = {115--129},
  doi       = {10.1007/978-3-031-25082-8_8}
}

@inproceedings{nagel2020up,
author = {Nagel, Markus and Amjad, Rana Ali and Van Baalen, Mart and Louizos, Christos and Blankevoort, Tijmen},
title = {Up or down? adaptive rounding for post-training quantization},
year = {2020},
publisher = {JMLR.org},
booktitle = {Proceedings of the 37th International Conference on Machine Learning},
articleno = {667},
numpages = {10},
series = {ICML'20}
}

@manual{hailo-8l,
    organization  = {{HAILO}},
    title         = {Hailo-8L M.2 AI Acceleration Module},
    number        = {},
    year          = {2024},
    month         = {3},    
    note          = {Rev. 1},
    url = {https://hailo.ai/files/hailo-8l-m-2-et-product-brief-en/},
}

@manual{orinnano,
    organization  = {{NVIDIA Corporation}},
    title         = {Jetson Orin Nano Developer Kit Carrier Board},
    number        = {},
    year          = {2024},
    month         = {4},    
    note          = {SP-11324-001 v1.2},
    url = {},
}

@software{custom-onnxruntime,
  author = {{Microsoft} and Piechocki, Mateusz},
  title = {{ONNX Runtime On-Device Training with HailoRT support. 1.22.1-hailo}},
  url = {https://github.com/MatPiech/onnxruntime},
  version = {1.22.1-hailo},
  date = {31-05-2026},
  note = {[Online]. Accessed: 31-05-2026}
}

@software{hailo,
  author = {{HAILO}},
  title = {{HailoRT. 4.23.0}},
  url = {https://hailo.ai},
  version = {4.23.0},
  date = {31-05-2026},
  note = {[Online]. Accessed: 31-05-2026}
}

@software{hailo-dataflow,
  author = {{HAILO}},
  title = {{Hailo Dataflow Compiler 3.31.0}},
  url = {https://hailo.ai},
  version = {3.31.0},
  date = {31-05-2026},
  note = {[Online]. Accessed: 31-05-2026}
}

\end{document}